\pdfoutput=1
\documentclass{article}

\usepackage{natbib}
\usepackage{amssymb}
\usepackage{amstext}
\usepackage{amsmath}
\usepackage{amsthm}
\usepackage{url}
\usepackage[ruled,linesnumbered]{algorithm2e}
\usepackage{graphicx}
\newcommand{\hyperplane}{\mathbf{w}}

\newcommand{\transpose}{^\textrm{T}}
\newcommand{\clabel}{y}
\newcommand{\ivector}{\mathbf{x}}
\newcommand{\learningf}{f}

\newcommand{\tsize}{m}
\newcommand{\fsize}{n}
\newcommand{\nzfsize}{s}
\newcommand{\xmatrix}{X}
\newcommand{\ymatrix}{\mathbf{y}}
\newcommand{\distribution}{D}

\newcommand{\espace}{\mathcal{Z}}
\newcommand{\examp}{z}
\newcommand{\tsequence}{Z}
\newcommand{\normalizer}{N}
\newcommand{\objective}{J}
\DeclareMathOperator*{\argmin}{arg\,min}
\newcommand{\erisk}{R_{emp}}
\newcommand{\subgradient}{\mathbf{a}}
\newcommand{\offset}{b}
\newcommand{\lfre}{c}
\newcommand{\sfre}{d}

\newcommand{\predicted}{\mathbf{p}}

\newtheorem{lemma}{Lemma}
\newtheorem{theorem}{Theorem}

\newtheorem{definition}{Definition}

\begin{document}

\title{Training linear ranking SVMs in linearithmic time using red-black trees}
\author{Antti Airola, Tapio Pahikkala and Tapio Salakoski}
\date{}

\maketitle

\begin{abstract}
We introduce an efficient method for training the linear ranking
support vector machine.
The method combines cutting plane optimization with
red-black tree based approach to subgradient calculations,
and has $O(\tsize\nzfsize+\tsize\log(\tsize))$ time complexity, where
$\tsize$ is the number of training examples, and $\nzfsize$ the average
number of non-zero features per example. Best previously known
training algorithms achieve the same efficiency only for
restricted special cases, whereas the proposed approach allows any real
valued utility scores in the training data. Experiments demonstrate the
superior scalability of the proposed approach, when compared to the fastest
existing RankSVM implementations.

\end{abstract}

\section{Introduction}

Learning to rank has been a task of significant interest during the recent
years. The ranking problem has been largely motivated by applications in
areas such as web search and recommender systems. Due to the large amounts
of data available in these domains, it is necessary for the used algorithms
to scale well, preferably close to linear time methods are needed.
For a detailed introduction to the topic of learning to rank, we refer to
\citep{liu2009learning,furnkranz2011preference}.

In this work we assume the so-called scoring setting, where
each data instance is associated with a utility score reflecting its goodness
with respect to the ranking criterion. A successful approach for learning ranking
functions has been to consider pairwise preferences
\citep{furnkranz2005preference}. In this setting, the aim is to minimize
the number of pairwise mis-orderings in the ranking produced when
ordering a set of examples according to predicted utility scores.
A number of machine learning algorithms optimizing relaxations of this
criterion have been proposed, such as the RankBoost \citep{freund2003rankboost},
RankNet \citep{burges05ranknet}, RankRLS
\citep{pahikkala2007rankrls,pahikkala2009ranking}, and the subject of this study,
the ranking support vector machine (RankSVM)
algorithm \citep{herbrich1999ordinal,joachims2002kddclickdata}.

The original solution proposed for RankSVM optimization was to train
a support vector machine (SVM) classifier on pairs of data examples.
Adapting standard dual SVM solvers to RankSVM training leads to 
$O(\tsize^4)$ scaling or worse, with respect to the training set size
$\tsize$ \citep{bottou2007svmsolvers}. While
linear RankSVMs can be trained more efficiently by solving the
primal optimization problem for SVMs
\citep{chapelle2007training,chapelle2009efficient},
still the complexity of any method that is explicitly trained on all the
pairs has at the very least quadratic dependence on the number of training
examples.

\citet{joachims2005support,joachims2006training} has shown
that linear RankSVM can be trained using cutting plane optimization,
also known as bundle optimization, much more efficiently in certain special
settings. \citet{joachims2005support} has proposed an
$O(\tsize\nzfsize+\tsize\log(\tsize))$ time algorithm,
where $\tsize$ is the number of training examples, and $\nzfsize$ the average
number of non-zero features per example, for the bipartite ranking
problem, where only two utility levels are allowed. The bipartite ranking
problem corresponds to maximizing the area under the receiving operating
characteristic curve (AUC) \citep{hanley82auc}, a performance measure 
widely used in machine learning (see e.g.
\citep{bradley1997,provost1998caseagainst}).
An $O(\tsize\nzfsize+\tsize\log(\tsize)+r\tsize)$
time generalization of the method has been presented for the case, where
$r$ different utility levels are allowed \citep{joachims2006training}.
\citet{chapelle2009efficient} have recently further
explored efficient methods for training RankSVM, the proposed methods
have similar scaling.

If $r$ is assumed to be a small constant, the existing methods are
computationally efficient. This is for example the case in the bipartite
ranking case where there are only two utility levels, corresponding to
the ``good'' and the ``bad'' objects. Similarly, movie ratings ranging
from one to five stars could be encoded using $r=5$ distinct utility levels.
However, in the general case where unrestricted scores are allowed, 
if most of the training examples have different scores $r\approx\tsize$
leading to $O(\tsize\nzfsize+\tsize^2)$ complexity. This worst scale quadratic
scaling with respect to the training set size limits the applicability
of RankSVM to large scale learning.

In this work we generalize the work of \citet{joachims2006training} and
present an $O(\tsize\nzfsize+\tsize\log(\tsize))$ time training
algorithm for linear RankSVM, that is applicable in the most general case, where
arbitrary real-valued utility scores are allowed in the training data. 
The method is based on
using modified red-black tree data structures \citep{bayer1972symmetric,
cormen2001introduction} for speeding up the 
evaluations needed in the optimization process.
Computational experiments show that the method has excellent scalability
properties also in practice.

In Section~\ref{learningsection} we introduce the general learning to rank
setting, and in Section~\ref{bmrmsection} we formalize the regularized
risk minimization problem, and present the general optimization framework 
for solving it. In Section~\ref{algorithmsection} 
we present our main contribution, the efficient RankSVM subgradient and loss
computation algorithms, and in Section~\ref{experimentsection} we present an
experimental evaluation of the
resulting training algorithm. We conclude in Section~\ref{conclusionsection}.


\section{Learning setting}\label{learningsection}

Let $\distribution$ be a probability distribution over a sample
space $\espace=\mathbb{R}^\fsize\times\mathbb{R}$. An example
$\examp=(\ivector,\clabel)\in\espace$ is a pair consisting
of an $\fsize$-dimensional column vector of real-valued features,
and an associated real-valued utility score.
Let the sequence
$\tsequence=((\ivector_1,\clabel_1),\ldots,(\ivector_\tsize,\clabel_\tsize))\in\espace^\tsize$
drawn according to $\distribution$ be a training set of $\tsize$ training
examples. $\xmatrix\in\mathbb{R}^{\fsize\times\tsize}$ denotes
the $\fsize\times\tsize$ data matrix whose columns contain the
feature representations of the training examples, and
$\ymatrix\in\mathbb{R}^{\tsize}$ is a column vector containing the utility
scores in the training set.
Our task is to learn from the training data a ranking function
$\learningf:\mathbb{R}^\fsize\rightarrow\mathbb{R}$. In the
linear case such a function can be
represented as $\learningf(\ivector)=\hyperplane\transpose\ivector$,
where $\hyperplane\in\mathbb{R}^\fsize$ is a vector of parameters.

The difference between ranking and regression is that in ranking,
the actual values taken by the prediction function are typically
not of interest.
Rather, what is of interest is how well the ordering
acquired by sorting a set of new examples according to their predicted scores matches
the true underlying ranking. This is a reasonable criterion for example in
the web search engines and recommender systems, where the task is to choose
a suitable order in which to present web pages or products to the end user.
A popular way to model this criterion is by considering the pairwise preferences
induced by a ranking (see e.g. \citep{furnkranz2005preference}). We say that an
example $\examp_i$ is preferred over example $\examp_j$, 
if $\clabel_i>\clabel_j$. In this case one would
require from the ranking function that $\learningf(\ivector_i)>\learningf(\ivector_j)$.

The performance of a ranking function can be measured by the pairwise ranking
error defined as
\begin{equation}\label{disagreement}
\frac{1}{\normalizer}\sum_{\clabel_i<\clabel_j}[\learningf(\ivector_i)>\learningf(\ivector_j)],
\end{equation}
where $\normalizer$ is the number of pairs for which $\clabel_i<\clabel_j$ holds
true. The equation (\ref{disagreement}) simply counts the number of swapped
pairs between the true ranking and the one produced by $\learningf$.

By restricting the allowed range of utility scores we can recover some popular
special cases of the introduced setting. In ordinal regression
\citep{herbrich1999ordinal,waegeman2008roc} it is assumed that there
exists a finite, often quite small set of possible
discrete ordinal labels. For example, movie ratings ranging from one star to
five stars would constitute such a scale. In the bipartite ranking task where
only two possible scores are allowed equation (\ref{disagreement}) becomes
equivalent to the Wilcoxon-Mann-Whitney formula used to calculate
AUC \citep{hanley82auc,cortes2003auc}.

In some learning to rank settings instead of having a total order over
all examples, the sample space is divided into disjoint subsets, and
pairwise preferences are induced only from pairwise comparisons
between the scores of examples in the same subset. An example of an
application settings where this approach is commonly adopted is document
retrieval, where data consists of query-document pairs, and the scores
represent the utility of the document with respect to the associated
user query \citep{joachims2002kddclickdata}. Preferences are induced only
between query-document pairs from the same query, never between examples from different queries.
In such settings we can calculate (\ref{disagreement}) separately
for each subset, and take the average value as the final error.

Minimizing (\ref{disagreement}) directly is computationally intractable,
successful approaches to learning to rank according to the pairwise criterion
typically minimize convex relaxations instead. The relaxation considered
in this work is the pairwise hinge loss, which together with a quadratic
regularizer forms the objective function of RankSVM. Before formally
defining the loss, we introduce a general optimization method suitable for
minimizing it.



\section{Bundle Methods for Regularized Risk Minimization}\label{bmrmsection}

A large class of machine learning algorithms can be formulated as the
unconstrained regularized risk minimization problem
\begin{equation}\label{regproblem}
\hyperplane^*=\argmin_{\hyperplane\in\mathbb{R}^\fsize}\objective(\hyperplane),
\end{equation}
where
\begin{equation*}
\objective(\hyperplane)=\erisk(\hyperplane)+\lambda\Arrowvert\hyperplane\Arrowvert^2,
\end{equation*}
$\hyperplane$ is the vector of parameters to be learned,
$\erisk$ is the empirical risk measuring how well $\hyperplane$
fits the training data, $\Arrowvert\hyperplane\Arrowvert^2$ is the
quadratic regularizer measuring the complexity of the considered
hypothesis, and $\lambda\in\mathbb{R}^+$ is a parameter. We assume
that $\erisk:\mathbb{R}^\fsize\rightarrow\mathbb{R}$ is convex
and non-negative.

Different choices of $\erisk$ result in different machine learning
methods such as SVM classification \citep{cortes1995support} or
regression \citep{drucker1997svmregression}, 
regularized least-squares regression \citep{poggio1990networks}, structured
output prediction methods \citep{tsochantaridis05structured}, 
RankRLS \citep{pahikkala2007rankrls,pahikkala2009ranking} and RankSVM
\citep{herbrich1999ordinal,joachims2002kddclickdata}.

Bundle methods for regularized risk minimization (BMRM)
\citep{teo2007scalable,smola07bundle,teo2010bundle}, is a general and
efficient optimization technique for solving (\ref{regproblem}).
The method is also known as the cutting plane method in the
machine learning literature, and it was under this name it was first
introduced for the purpose of efficiently optimizing large margin type
of loss functions \citep{tsochantaridis05structured}.
It was later shown, that the method can be generalized to
arbitrary convex loss functions, as long as subgradient evaluations for the
loss can be done efficiently \citep{teo2007scalable,smola07bundle}. In the
following we adopt our notation and terminology from the BMRM literature.

BMRM iteratively constructs a piecewise linear lower bound
approximation of $\erisk$. Let $R_t$ be the piecewise linear approximation
at iteration $t$. We approximate
(\ref{regproblem}) with
\begin{equation}\label{reduced_problem}
\hyperplane_t=\argmin_{\hyperplane\in\mathbb{R}^\fsize}\objective_t(\hyperplane),
\end{equation}
where
\begin{equation*}
\objective_t(\hyperplane)=R_t(\hyperplane)+\lambda\Arrowvert\hyperplane\Arrowvert^2.
\end{equation*}
Thus the regularizer remains the same, but the empirical risk term
is replaced with the piecewise linear lower bound.

The empirical risk is lower bounded by its first order Taylor
approximation at any $\hyperplane'\in\mathbb{R}^\fsize$, defined as
\begin{equation*}
\erisk(\hyperplane)\geq\erisk(\hyperplane')+\langle\hyperplane-\hyperplane',\subgradient'\rangle,
\end{equation*}
where $\subgradient'$ is any subgradient of $\erisk$ at $\hyperplane'$.
By defining
$\offset'=\erisk(\hyperplane')-\langle\hyperplane',\subgradient'\rangle$
due to the linearity of the inner product this can be re-written as
\begin{equation*}
\erisk(\hyperplane)\geq\langle\hyperplane,\subgradient'\rangle+\offset',
\end{equation*}
$\langle\hyperplane,\subgradient'\rangle+\offset'=0$ is called
a cutting plane.

Using several cutting planes BMRM approximates $\erisk$ with
\begin{equation*}
R_t(\hyperplane)=\max_{i=1\ldots t}
\{\langle\hyperplane,\subgradient_i\rangle+\offset_i\}\}.
\end{equation*}
Using this approximation (\ref{reduced_problem}) can be
solved by solving an equivalent quadratic program whose size depends
on the number of cutting planes used.

\begin{algorithm}\label{bmrm}
\KwIn{$\hyperplane_0$, $\epsilon\ge 0$}  
\KwOut{$\hyperplane_b$}
$t \leftarrow 0$\;
$\hyperplane_b\leftarrow\hyperplane_0$\;
\Repeat
{$\epsilon_t<\epsilon$}
{$t\leftarrow t+1$\;
$\subgradient_t\leftarrow$ subgradient of $\erisk$ at $\hyperplane_{t-1}$\;
$\offset_t\leftarrow\erisk(\hyperplane_{t-1})-\langle\hyperplane_{t-1},\subgradient_t\rangle$\;
Update $R_t(\hyperplane)$ by adding the new cutting plane
$\langle\cdot,\subgradient_t\rangle+\offset_t$\;
$\hyperplane_t\leftarrow\argmin_{\hyperplane}\objective_t(\hyperplane)$\;
\If{$\objective(\hyperplane_t)<\objective(\hyperplane_b)$}
{$\hyperplane_b\leftarrow\hyperplane_t$}
$\epsilon_t\leftarrow\objective(\hyperplane_b)-\objective_t(\hyperplane_t)$
}
\caption{BMRM}
\end{algorithm}

The procedure according to which BMRM builds the lower bound is described
in algorithm~\ref{bmrm}.
The formulation given here differs slightly from that of
\citet{teo2010bundle} in that following
the suggestion of \citet{franc2009optimized} we maintain the best
this far seen solution $\hyperplane_b$, and only update the solution
when the new solution $\hyperplane_t$ is better.
The parts of the algorithm which depend on the choice of $\erisk$
are the calculation of a subgradient and the value of
$\erisk$ at point $\hyperplane_{t-1}$, as well as the termination criterion.

The rate of convergence is for BMRM independent of the training set size
\citep{smola07bundle}.
The size of the quadratic program solved on line 8 does grow with
the number of iterations, but its computational cost is on large datasets
insignificant compared to the cost of computing the cutting plane.
Thus what is required for efficient learning with a convex loss function is
an efficient algorithm for computing its value and subgradient.

\section{Efficient computation of loss and subgradient}\label{algorithmsection}

Next we present an efficient algorithm for evaluating the empirical
risk and its subgradient for RankSVM. 
First, in Section~\ref{prelsection} we recall results from
\citet{joachims2006training} to identify the computational bottleneck in RankSVM training,
which occurs in the loss and subgradient computation.
Next, in Section~\ref{treesection} we introduce search tree algorithms,
which we use to speed up these computations. Building on these results
we present an efficient algorithm for loss and subgradient computation in
Section~\ref{subgradientsection}.

\subsection{Preliminaries}\label{prelsection}
The following results were first presented by
\citet{joachims2006training}, who formulated the RankSVM optimization problem as
a constrained optimization problem.
In this work we follow a different but equivalent formulation of
RankSVM as an unconstrained optimization problem within the BMRM framework
(see \citet{teo2010bundle}).

The average pairwise hinge loss computed over the training set, with respect to
a given solution $\hyperplane$ is
\begin{equation}\label{hinge_loss}
\erisk(\hyperplane)=
\frac{1}{\normalizer}\sum_{\clabel_i<\clabel_j}
\max(0,1+\hyperplane\transpose\ivector_i-\hyperplane\transpose\ivector_j),
\end{equation}
where $\normalizer$ is the number of pairs for which $\clabel_i<\clabel_j$ holds
true.

The pairwise hinge loss, together with the quadratic regularizer forms the
objective function of RankSVM. The most obvious approach to evaluating (\ref{hinge_loss}),
and also its subgradient involves going explicitly through all the
pairs in the training set. However, this would lead to $O(\tsize^2)$ complexity,
which is inefficient on large data sets. Let us define
\begin{equation}\label{cdefinition}
c_i=\arrowvert \{j : (\clabel_i < \clabel_j) \wedge
(\hyperplane\transpose\ivector_i>\hyperplane\transpose\ivector_j-1)
\wedge (1\leq j
\leq\tsize)\}\arrowvert
\end{equation}
and 
\begin{equation}\label{ddefinition}
d_i=\arrowvert \{j : (\clabel_i > \clabel_j) \wedge
(\hyperplane\transpose\ivector_i<\hyperplane\transpose\ivector_j+1)
\wedge (1\leq j
\leq\tsize)\}\arrowvert.
\end{equation}

Using the frequencies (\ref{cdefinition}) and (\ref{ddefinition}), we
recover an alternative formulation for the empirical risk.
\begin{lemma}[\citet{joachims2006training,teo2010bundle}]\label{lem:loss}
The average pairwise hinge loss (\ref{hinge_loss}) can be equivalently
expressed as
\begin{equation}\label{ranksvm_loss}
\frac{1}{\normalizer}\sum_{i=1}^{\tsize}((\lfre_i-\sfre_i)\hyperplane\transpose\ivector_i+\lfre_i).
\end{equation}
 \end{lemma}

Similarly, computation of (\ref{cdefinition}) and (\ref{ddefinition}) allows the
computation of a subgradient of the empirical risk.
 \begin{lemma}[\citet{joachims2006training,teo2010bundle}]\label{lem:subgradient}
A subgradient of (\ref{hinge_loss}) can be expressed as
\begin{equation}\label{ranksvm_grad}
\nabla\erisk(\hyperplane)=\frac{1}{\normalizer}\sum_{i=1}^{\tsize}
(\lfre_i-\sfre_i)\ivector_i.
\end{equation}
\end{lemma}

Inner product evaluations, scalar-vector multiplications
and vector summations are needed to calculate
(\ref{ranksvm_loss}) and (\ref{ranksvm_grad}). These take each
$O(\nzfsize)$ time, the average number of nonzero elements in a
feature vector. Provided that we know the values of $\lfre_i$,
$\sfre_i$ and $\normalizer$, both the loss and subgradient can thus
be evaluated in $O(\tsize\nzfsize)$ time.

\citet{joachims2006training} (and equivalently \citet{teo2010bundle}) describe a
way to calculate efficiently these frequencies, and subsequently the loss and
subgradient. However, the work assumes that the range of possible utility
score values is restricted to $r$ different values, with
$r$ being quite small. The algorithm requires $O(r)$ passes through the
training set, contributing a $O(r\tsize)$ term to the overall complexity, which
is  $O(\tsize\nzfsize+\tsize\log(\tsize)+r\tsize)$.
If the number of allowed scores is not restricted, at worst case $r=\tsize$ with
the resulting complexity $O(\tsize\nzfsize+\tsize^2)$, meaning quadratic
behavior with respect to the training set size. However, as we show
next, the dependence on $r$ can be removed from the algorithm by utilizing
order statistics trees, resulting in
$O(\tsize\nzfsize+\tsize\log(\tsize))$ cost also in
the most general case, where arbitrary real valued utility scores are allowed.

\subsection{Order statistics tree}\label{treesection}

 \newcommand{\tree}{T}
 \newcommand{\tinsert}{\textnormal{Tree-Insert}}
 \newcommand{\csmaller}{\textnormal{Count-Smaller}}
 \newcommand{\clarger}{\textnormal{Count-Larger}}
 \newcommand{\newnode}{x}
 \newcommand{\newkey}{k}
 \newcommand{\tempnode}{y}
 \newcommand{\secondnode}{z}
 \newcommand{\nodeleft}{\textnormal{left}}
 \newcommand{\noderight}{\textnormal{right}}
 \newcommand{\noderoot}{\textnormal{root}}
 \newcommand{\nodeparent}{\textnormal{par}}
 \newcommand{\nodechild}{\textnormal{child}}
 \newcommand{\nodekey}{\textnormal{key}}
 \newcommand{\nodesize}{\textnormal{nodesize}}
 \newcommand{\treesize}{\textnormal{size}}

The order statistics tree \citep{cormen2001introduction} is a balanced
binary search tree, which has been augmented to support logarithmic
time computation of order statistics, such as the rank of a given element
in the tree, or the recovery of the $k$:th element in the tree. As we
will show in the following, the data structure also allows efficient
computation of the frequencies needed in the RankSVM loss and subgradient
computations. Next, we introduce the order statistics tree, recall its
main properties, and introduce algorithms which act as building blocks
for the fast RankSVM training method. In the following, we assume that
the number of elements inserted to the search tree is bounded
by $\tsize$.

The binary search tree is one of the most fundamental data structures in
computer science. It is a linked data structure consisting of nodes. A given
node $\newnode$ contains a real valued $\nodekey(\newnode)$, and
pointers to a parent node $\nodeparent(\newnode)$, a left child
$\nodeleft(\newnode)$, and a right child $\noderight(\newnode)$. A parent
or a child may be missing, denoted by the value $\emptyset$. Only a
single node known as the root node is allowed to have missing parent.
Nodes with no children are called leaf nodes. The root of an empty tree is
assumed to be $\emptyset$. The height of a binary search tree is the length
of the path from the root node to the lowest leaf node, and the size of a
tree is the number of elements it contains.
We call the tree rooted at $\nodeleft(\newnode)$
the left subtree of $\newnode$, and the tree rooted at $\noderight(\newnode)$
the right subtree of $\newnode$. 
The tree always satisfies the \emph{binary search tree property}, requiring
that the nodes stored in the left subtree of a node have smaller than or equal
keys as the node, and the nodes in the right subtree have larger than
or equal keys as the node.

The worst-case cost of searching for an element in the tree, or inserting or
deleting an element is proportional to the height of the tree. To guarantee
the efficiency of such operations, self-balancing binary trees combine basic
tree update operations with maintenance operations that maintain
$O(\log(\tsize))$ tree height. One of the most popular self-balancing search
tree variants is the red-black tree
\citep{bayer1972symmetric,cormen2001introduction}.
Further, \citep{cormen2001introduction} describe a modified variant of the red-black tree
defined as follows:

\begin{definition}[Order statistics
tree (\citet{cormen2001introduction})]\label{ostree}
The order statistics tree is a self-balancing binary search tree, with
the following properties
\begin{itemize}
  \item The binary search tree property:
``Let $\newnode$ be a node in a binary search tree. If $\tempnode$ is a node
in the left subtree of $\newnode$, then
$\nodekey(\tempnode)\leq\nodekey(\newnode)$. If $\tempnode$ is a node in the
right subtree of $\newnode$, then $\nodekey(\newnode)\leq\nodekey(\tempnode)$.''
\item Balance: $O(\log(\tsize))$ height after arbitrary
insertions and deletions.
\item Each node of the order statistics tree stores an additional attribute
$\treesize$ defined as
\begin{equation*}
\treesize(\newnode)=\left\{\begin{array}{ll}
0 &\textrm{ if } \newnode=\emptyset\\
\treesize(\nodeleft(\newnode))+ \treesize(\noderight(\newnode))+1 &\textrm{
otherwise }\\
 \end{array}\right..
 \end{equation*}
The correct value for this attribute is maintained after arbitrary insertions
and deletions.
\end{itemize}

\end{definition}
Note that the definition allows the existence of multiple nodes with
the same key value in the tree.

Let $\tinsert(\tree,\newnode)$ be the insertion operation of an arbitrary node
$\newnode$ to an order statistics tree $\tree$, whose size is bounded by
$\tsize$. $\tinsert$ adds the new node to the tree, maintaining the binary
search tree property, the correct values of the $\treesize$ attribute in the
nodes, and the $O(\log(\tsize))$ height of the tree. The time complexity
of the operation is characterized by the following Lemma:

 \begin{lemma}[\citet{cormen2001introduction}]\label{lem:insert} The time
 complexity of $\tinsert$ to order statistics tree is $O(\log(\tsize))$.
 \end{lemma}

Further, for the RankSVM computations we require routines that efficiently
compute the number of elements in the tree with a smaller, or a larger value
than a given argument. Let $\newkey$ be a real value, and let $\tree$ be
an order statistics tree, whose size is bounded by $\tsize$. Then,
$\csmaller(\noderoot(\tree),\newkey)$ (Algorithm~\ref{alg:csmaller}) returns the
number of nodes in $\tree$ with a smaller key value than $\newkey$.
The algorithm $\clarger(\noderoot(\tree),\newkey)$, which is not presented
separately, works in analogous fashion returning the number of nodes with
a larger key value than the argument.

 \begin{lemma}\label{lem:countcorrect} The correctness
 of $\csmaller$ and $\clarger$.
 \end{lemma}
 \begin{proof}
Let $C(\newnode,\newkey)$ denote the number of values smaller than $\newkey$
stored in a binary search tree whose root $\newnode$ is. 
Due to Definition~\ref{ostree}, the following always holds
\begin{equation*}
C(\newnode,\newkey)=\left\{\begin{array}{ll}
0 &\textrm{ if } \newnode=\emptyset\\
C(\noderight(\newnode),\newkey)+\treesize(\nodeleft(\newnode))+1
&\textrm{ if } \nodekey(\newnode)<\newkey\\
C(\nodeleft(\newnode),\newkey)&\textrm{ otherwise }\\
\end{array}\right..
\end{equation*}
This recursive equation supplies us directly with an algorithm for
computing $C(\newnode,\newkey)$, which is implemented in $\csmaller$.
The proof for $\clarger$ is analogous.
 \end{proof}
 
 \begin{lemma}\label{lem:count}
 Count-Smaller and Count-Larger have $O(\log(\tsize))$ complexity.
 \end{lemma}
 \begin{proof}
 On each
 call of $\csmaller$ or $\clarger$, $O(1)$ cost computations are performed, and
 additionally the routine may call itself once with a child node of the input node.
 At worst case the recursion proceeds until the lowest leaf node in the tree is
 reached, requiring a number of calls proportional to the height of the tree,
 which is according to Definition~\ref{ostree} guaranteed to be of the order
 $O(\log(\tsize))$.
 \end{proof}

\begin{algorithm}\label{alg:csmaller}
\KwIn{$\newnode$, $\newkey$}
\lIf{$\newnode=\emptyset$}
{
\Return 0\;}
\lElseIf{$\nodekey(\newnode) < \newkey$}
{\Return
$\csmaller(\noderight(\newnode),\newkey)+\treesize(\nodeleft(\newnode))+1$}\;
\Return $\csmaller(\nodeleft(\newnode),\newkey)$\;
\caption{$\csmaller$}
\end{algorithm}


Let $r$ be the number of distinct keys stored in the tree. In the presence of a
large number of duplicates, meaning $r << \tsize$, the order statistics tree can
be implemented more efficiently by storing duplicate keys to the same
node. In such an implementation, each node contains an additional attribute
$\nodesize(\newnode)$, which measures how many times
$\nodekey(\newnode)$ has been inserted to the tree. When inserting a new key,
a new node is not created if the key already exists in the tree,
but rather the $\nodesize$ attribute of the existing node is incremented
by one. Thus, we need to
re-define $\treesize(\newnode)=\treesize(\nodeleft(\newnode))+
\treesize(\noderight(\newnode))+\nodesize(\newnode)$. For this modified
variant of the order statistics tree, the height of the tree, and the cost of
$\tinsert$, $\csmaller$ and $\clarger$
are bounded by $O(\log(r))$.
However, this improvement does not translate into further improvements in the asymptotic cost of RankSVM
training, due to $O(\tsize\log(\tsize))$ cost of a sorting operation that is
required on each iteration of training.

\subsection{Subgradient computation}\label{subgradientsection}

\begin{algorithm}\label{ranksvm}
\KwIn{$\xmatrix$, $\ymatrix$, $\hyperplane$, $\normalizer$}  
\KwOut{$\subgradient$, loss}
$\predicted\leftarrow\xmatrix\transpose\hyperplane$\;
$\mathbf{\lfre}\leftarrow$ $\tsize$ length column vector of zeros\;
$\mathbf{\sfre}\leftarrow$ $\tsize$ length column vector of zeros\;
$\pi\leftarrow$ training set indices, sorted in ascending
order according to $\predicted$\;
$\tree\leftarrow$ new empty search tree\;
$j\leftarrow 1$\;
\For{$i\leftarrow 1$ \KwTo $\tsize$}
{
\While{$(j\leq\tsize)$ and ($\predicted[\pi[i]]]>\predicted[\pi[j]]-1)$}
{$\tinsert(\tree,\ymatrix[\pi[j]])$\; $j\leftarrow j+1$\;
}
$\mathbf{\lfre}[\pi[i]]\leftarrow\clarger(\noderoot(\tree),\ymatrix[\pi[i]])$\;
}
$\tree\leftarrow$ new empty search tree\;
$j\leftarrow \tsize$\;
\For{$i\leftarrow \tsize$ \KwTo $1$}
{
\While{$(j\geq 1)$ and ($\predicted[\pi[i]]<\predicted[\pi[j]]+1)$}
{$\tinsert(\tree,\ymatrix[\pi[j]])$\;
$j\leftarrow j-1$\;
}
$\mathbf{\sfre}[\pi[i]]\leftarrow\csmaller(\noderoot(\tree),\ymatrix[\pi[i]])$\;
}
loss$\leftarrow\frac{1}{\normalizer}\sum_{i=1}^{\tsize}((\mathbf{\lfre}[i]-\mathbf{\sfre}[i])*\predicted[i]+\mathbf{\lfre}[i])$\;
$\subgradient\leftarrow\frac{1}{\normalizer}\xmatrix(\mathbf{\lfre}-\mathbf{\sfre})$\;
\caption{Subgradient and loss computation}
\end{algorithm}

Algorithm~\ref{ranksvm} presents the main contribution of this paper,
an $O(\tsize\nzfsize+\tsize\log(\tsize))$
time method for calculating the RankSVM loss and subgradient. The algorithm
uses order statistics trees to efficiently compute the frequencies 
(\ref{cdefinition}) and (\ref{ddefinition}), which are necessary in
computing the value of the loss (\ref{ranksvm_loss}) and a subgradient
(\ref{ranksvm_grad}).

\begin{theorem}\label{theo:subgrad1}
The correctness of Algorithm~\ref{ranksvm}.
\end{theorem}
\begin{proof}

The algorithm computes the predicted utility scores for
the training set with $\predicted=\xmatrix\transpose\hyperplane$
(note that $\predicted[i] = \hyperplane\transpose\ivector_i$).
Next, an index list $\pi$ is created, where
the indices of the training examples are ordered in an increasing
order, according to the magnitudes of their predicted scores, so that
$\predicted[\pi[1]]\leq\predicted[\pi[2]]\leq\ldots\leq\predicted[\pi[\tsize]]$.

Let us consider the $i$:th iteration of the for-loop on lines $7-11$,
with $1\leq i \leq\tsize$. After
the while loop on lines $8-10$ has been executed,
the index $j$ divides the training set into two parts.
For $1\leq k< j$ it holds that
$\predicted[\pi[i]]>\predicted[\pi[k]]-1$ and for $j\leq
k\leq\tsize$ it holds that $\predicted[\pi[i]]\leq\predicted[\pi[k]]-1$.
The keys corresponding to the indices $\pi[1]\ldots\pi[j-1]$ have,
either on this or on previous iterations, been inserted to the order
statistics tree $\tree$, by the $\tinsert$ call on line $9$. Therefore, on line
$11$ where $\clarger(\noderoot(\tree),\ymatrix[\pi[i]])$ is
called, $\tree$ contains the labels of the training examples indexed by the
set
\begin{equation*}
\{ k  : (\predicted[\pi[i]]>\predicted[k]-1) \wedge (1\leq k
\leq\tsize)\}
\end{equation*}
According to Lemma~\ref{lem:count},
$\clarger(\noderoot(\tree),\ymatrix[\pi[i]])$ returns the number of keys in
$\tree$, with a larger value than the argument $\ymatrix[\pi[i]]$. Thus, line
$11$ stores to $\mathbf{\lfre}[\pi[i]]$ the value
\begin{equation*}
\arrowvert\{ k  :
(\ymatrix[\pi[i]]<y_k)\wedge(\predicted[\pi[i]]>\predicted[k]-1) \wedge (1\leq
k \leq\tsize)\}\arrowvert,
\end{equation*}
which, according to (\ref{cdefinition}) is the frequency $c_{\pi[i]}$.

It can be verified analogously, that the for-loop on lines $14-18$ computes
the frequencies $d_1\ldots d_\tsize$ according to (\ref{ddefinition}).
After line $18$ has been
executed, the algorithm has thus filled two arrays,
$\mathbf{\lfre}=[c_1, \ldots, c_\tsize]$ and $\mathbf{\sfre}=[d_1, \ldots,
d_\tsize]$. Using these frequencies, the loss is computed on line $19$
according to (\ref{ranksvm_loss}), and the subgradient on line $20$
according to (\ref{ranksvm_grad}).

\end{proof}


\begin{theorem}\label{theo:subgrad}
The complexity of calculating the
loss and the subgradient with Algorithm~\ref{ranksvm} is 
$O(\tsize\nzfsize+\tsize\log(\tsize))$ for any training set of size $\tsize$ and
sparsity $\nzfsize$, with unrestricted range for utility
score values allowed.
\end{theorem}
\begin{proof}
The cost of computing $\xmatrix\transpose\hyperplane$  on line $1$ using
standard sparse matrix - vector product multiplication algorithms  is
$O(\tsize\nzfsize)$. Initializing the m-length arrays on lines $2$ and
$3$ is a $O(\tsize)$ operation, whereas the empty search tree initializations
in lines $5$ and $12$ take $O(1)$ time. The sorting operation on line
$4$ can be done in $O(\tsize\log(\tsize))$ using for example the
heapsort algorithm \citep{williams1964heapsort}. $\tinsert$ on
line $9$, and $\clarger$ on line $11$ are both called exactly $\tsize$ times,
once for each training example. According to Lemmas \ref{lem:insert} and
\ref{lem:count}, both operations have $O(\log(\tsize))$ cost, resulting in
$O(\tsize\log(\tsize))$ complexity for the lines $7-11$ altogether.
Analogously, lines $14-18$ have also $O(\tsize\log(\tsize))$ cost,
since $\tinsert$ on line $16$ and the $\csmaller$ on line $18$
are both called exactly $\tsize$ times. Finally, the
loss computation on line $19$ requires $O(\tsize)$ floating point operations,
and the subgradient computation on line $20$ requires a $O(\tsize\nzfsize)$
matrix-vector product. Summing all the complexities together,
the resulting computational complexity of Algorithm~\ref{ranksvm} is
$O(\tsize\nzfsize+\tsize\log(\tsize))$.

\end{proof}

Next, we present a theorem that characterizes the overall complexity of
RankSVM training using the introduced subgradient and loss computation
algorithm. The theorem and its proof are similar to those presented by
\citet{joachims2006training}. However, unlike Joachims, we do not
(implicitly) assume the number of different relevance level to be
constant.

\begin{theorem}
For any fixed $\epsilon>0$ and $\lambda>0$,
the computational cost of linear RankSVM training with BMRM
(Algorithm~\ref{bmrm}) using Algorithm~\ref{ranksvm} for loss and
subgradient computations is $O(\tsize\nzfsize+\tsize\log(\tsize))$ for any
training set of size $\tsize$ and sparsity $\nzfsize$, with unrestricted range for utility
score values allowed.
\end{theorem}

\begin{proof}
Theorem~5 in \citep{smola07bundle} states that the BMRM has, under minor
technical assumptions, $O(\frac{1}{\epsilon\lambda})$ speed of convergence
to $\epsilon$-accurate solution. The convergence speed
does not depend on the values of $\tsize$ and $\nzfsize$.
During initialization, the exact value of $\normalizer$
can be computed in $O(\tsize\log(\tsize))$ by sorting the true
utility scores of the training examples.
Further, the only computations within each iteration
that depend on  $\tsize$ and $\nzfsize$ are the loss and the subgradient
computations. Therefore, the computational complexity of training RankSVM
is the same as that of Algorithm \ref{ranksvm}, which is according to Theorem
\ref{theo:subgrad} $O(\tsize\nzfsize+\tsize\log(\tsize))$.
\end{proof}

As discussed previously, in some ranking settings we do not have a global
ranking over all examples. Instead, the training data may be divided into
separate subsets, over each of which a ranking is defined. Let the training
data set be divided into $R$ subsets, each consisting on average of
$\frac{\tsize}{R}$ examples. Then we can calculate the loss and the subgradient
as the average over the losses and subgradients for each subset. The
computational complexity becomes
$O(R*(\frac{\tsize}{R}\nzfsize+\frac{\tsize}{R}\log(\frac{\tsize}{R}))
= O(\tsize\nzfsize+\tsize\log(\frac{\tsize}{R}))$.

\section{Computational experiments}\label{experimentsection}

\begin{figure}[t]
\centering
\begin{tabular}{cc}
\includegraphics[width=0.50\linewidth]{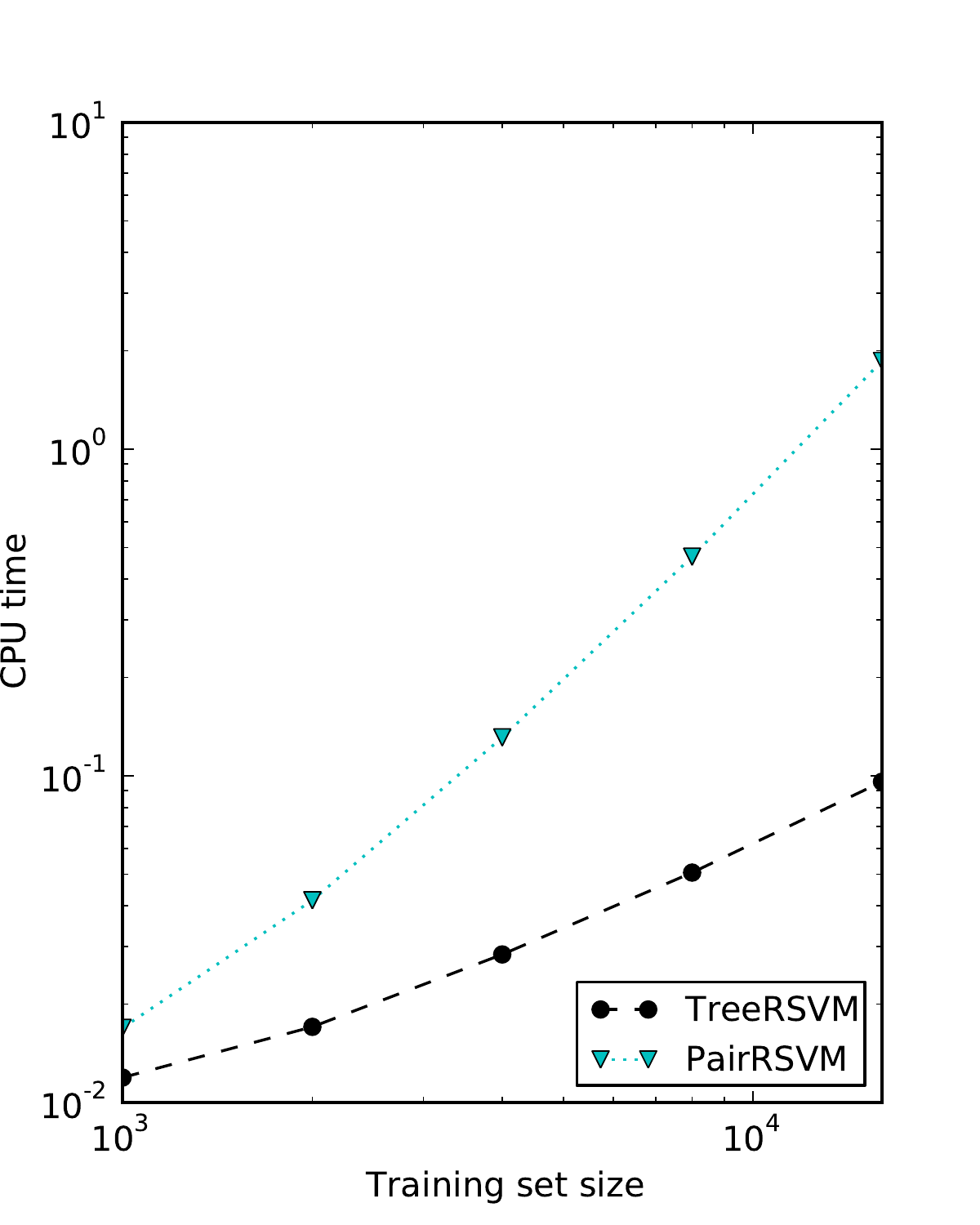} &
\includegraphics[width=0.50\linewidth]{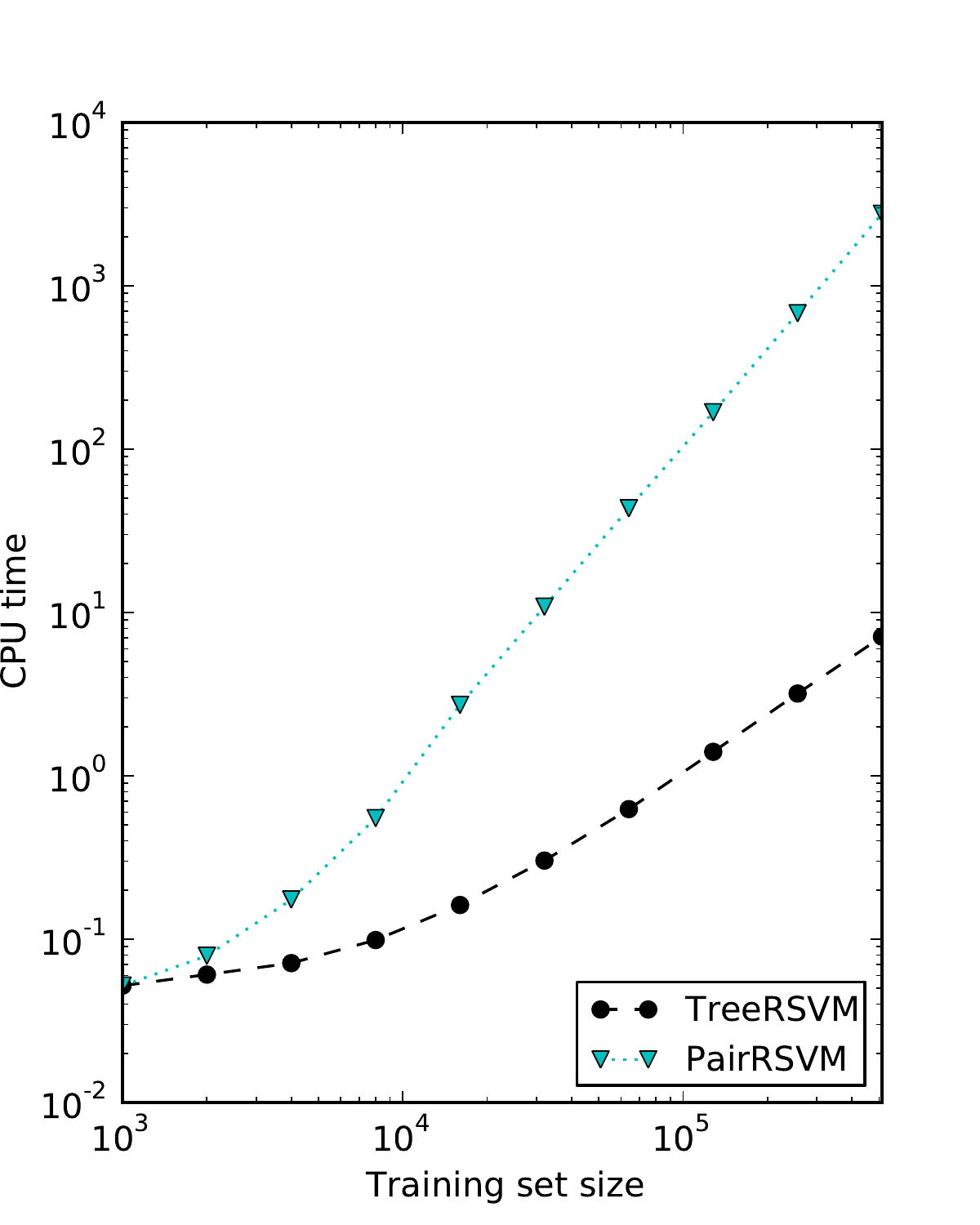}\\
\end{tabular}
\caption{Average iteration cost. Cadata (left) and Reuters (right).}
\label{fig:itcost}
\end{figure}

\begin{figure}[t]
\centering
\begin{tabular}{cc}
\includegraphics[width=0.50\linewidth]{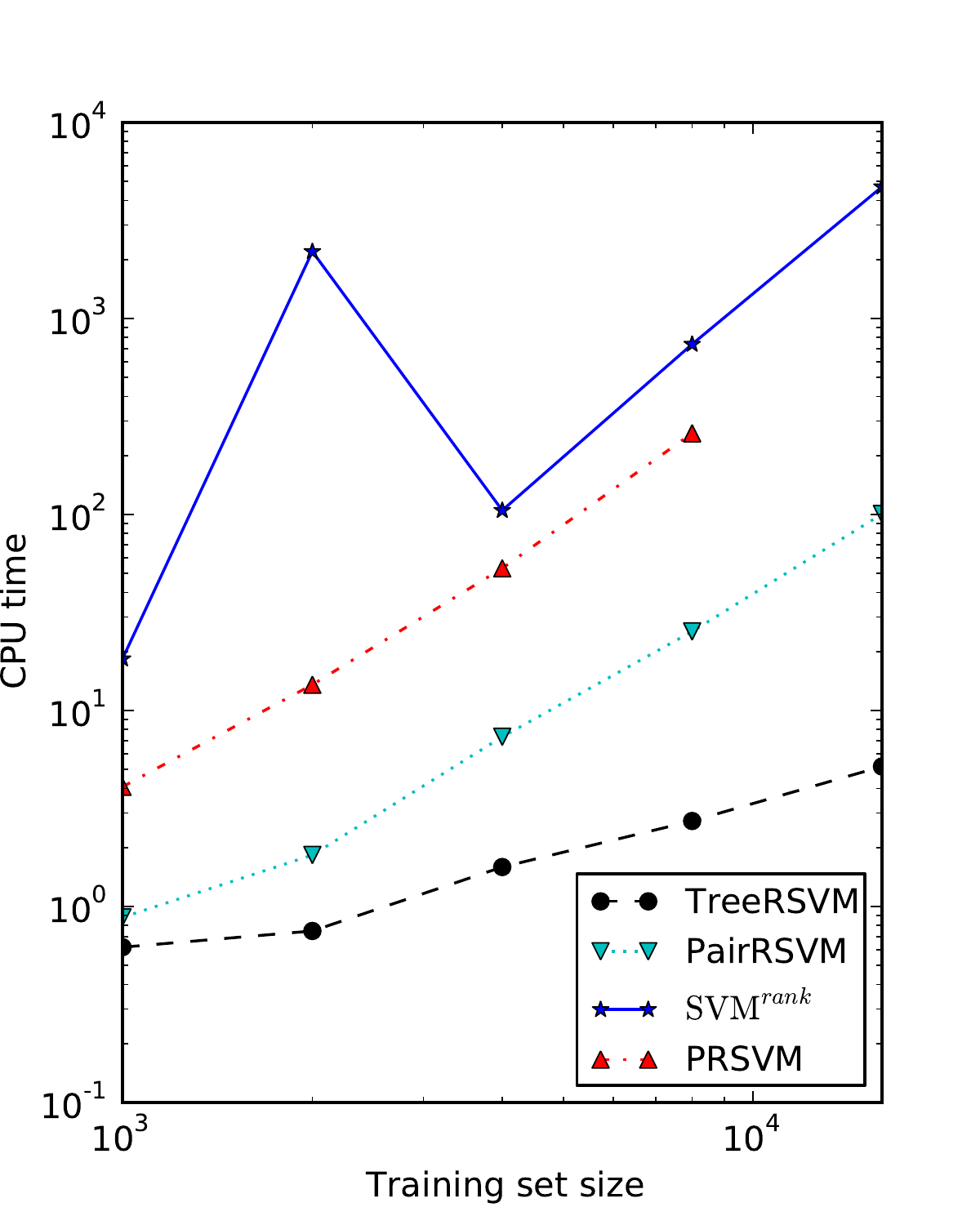} &
\includegraphics[width=0.50\linewidth]{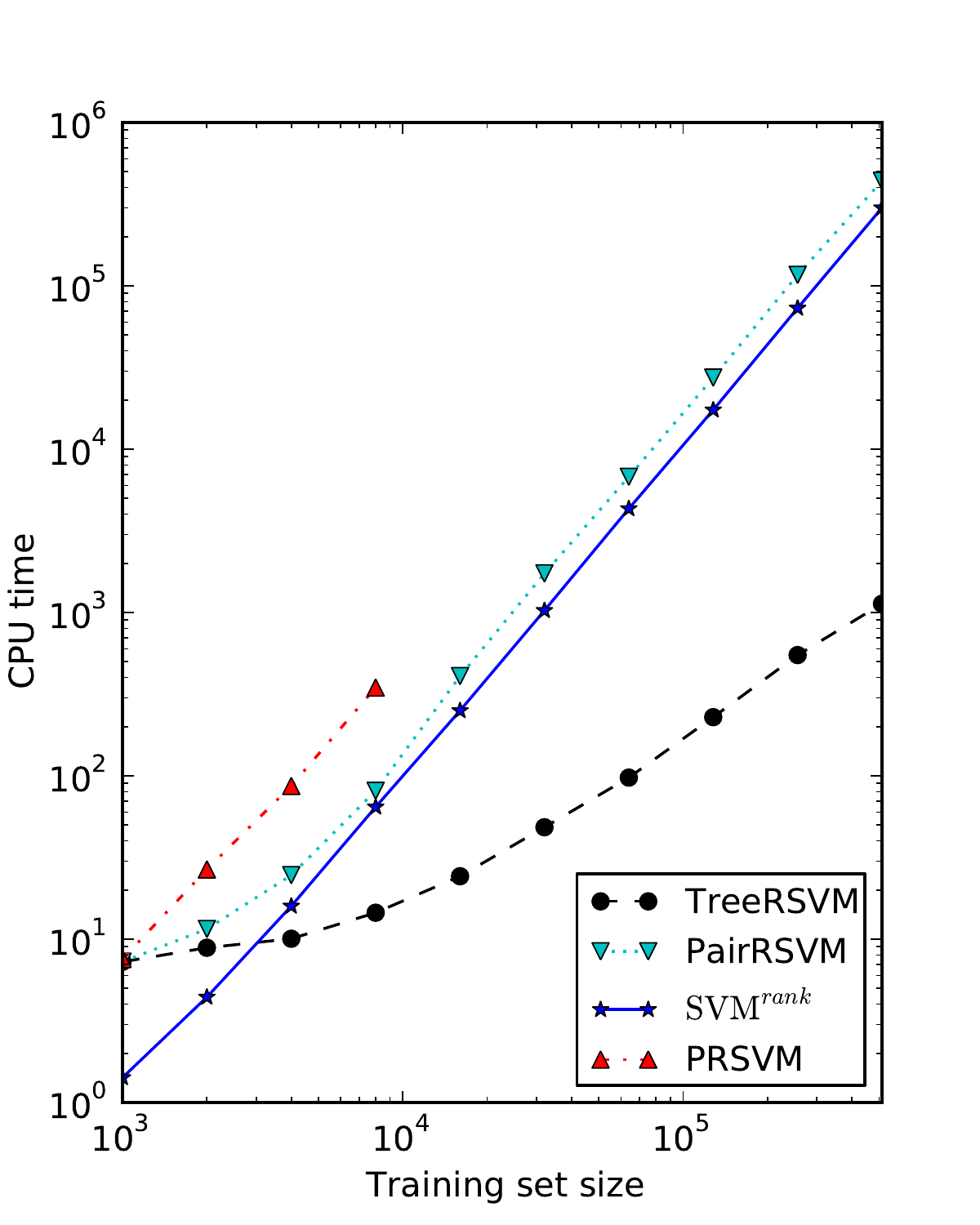}\\
\end{tabular}
\caption{Runtimes for different RankSVM implementations. Cadata (left) and
Reuters (right).}
\label{fig:rtimes}
\end{figure}

\begin{figure}[t]
\centering
\begin{tabular}{c}
\includegraphics[width=0.80\linewidth]{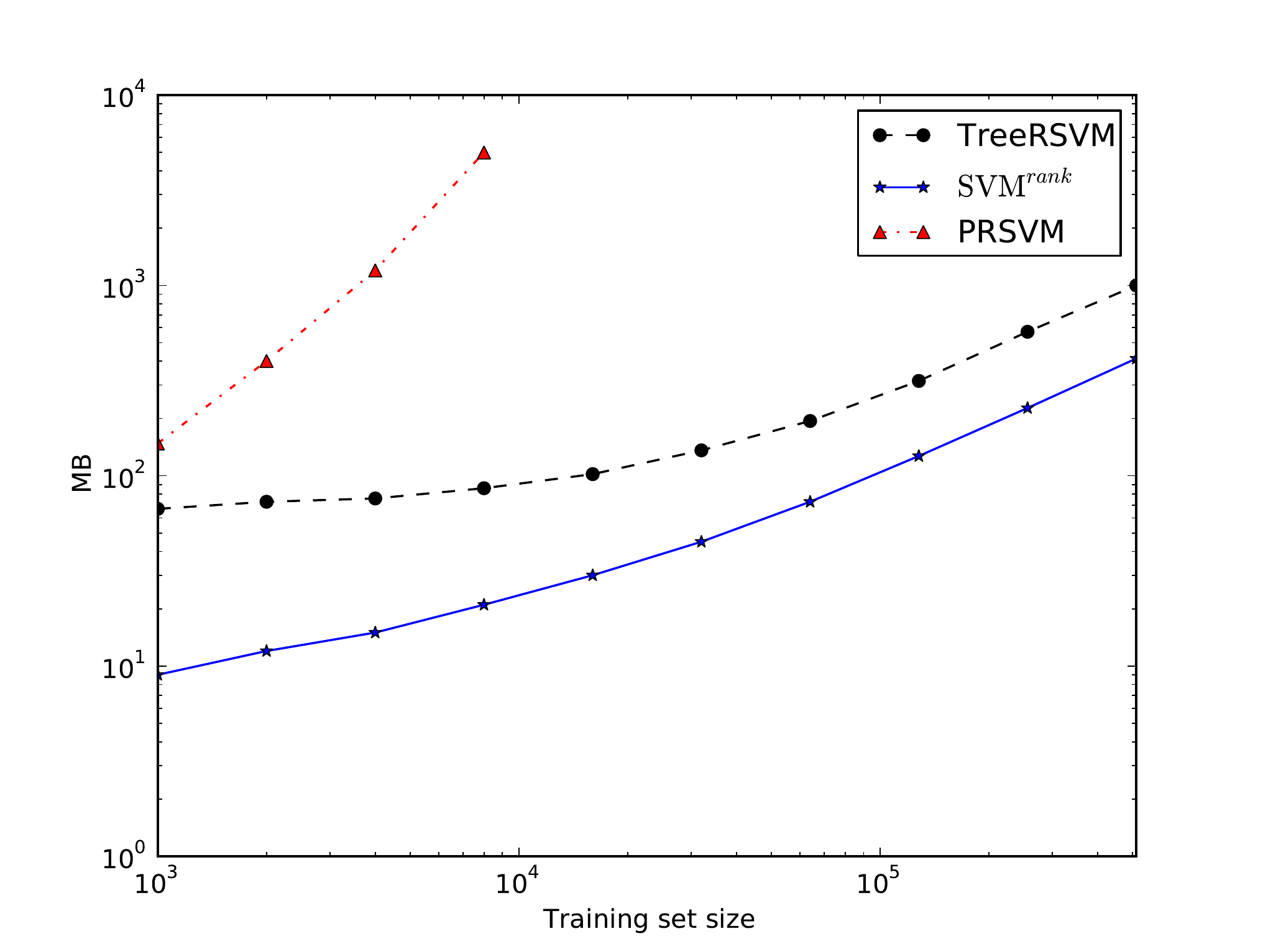}\\
\end{tabular}
\caption{Memory usage for different RankSVM implementations, measured on
Reuters.}
\label{fig:memuse}
\end{figure}

\begin{figure}[t]
\centering
\begin{tabular}{cc}
\includegraphics[width=0.50\linewidth]{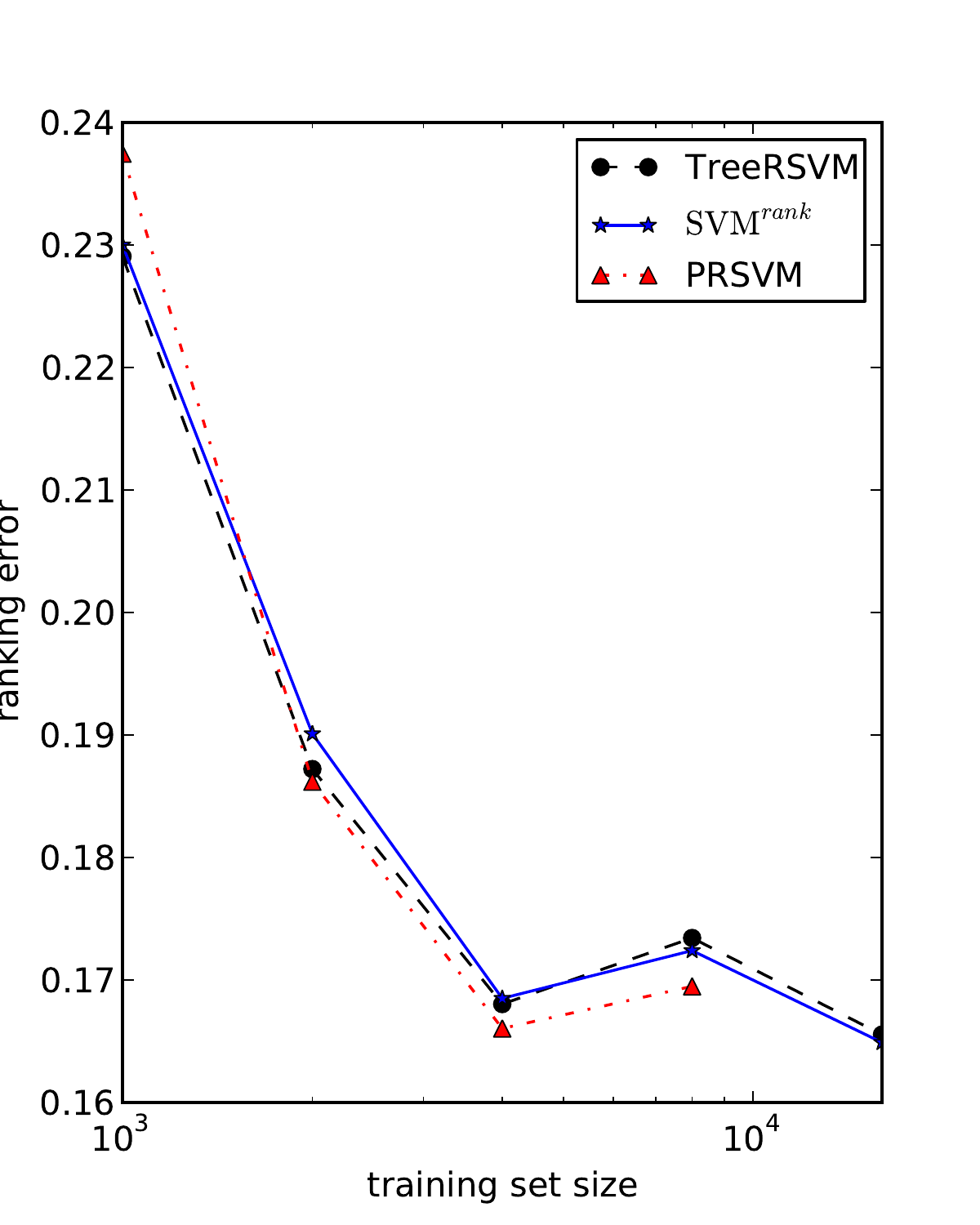} &
\includegraphics[width=0.50\linewidth]{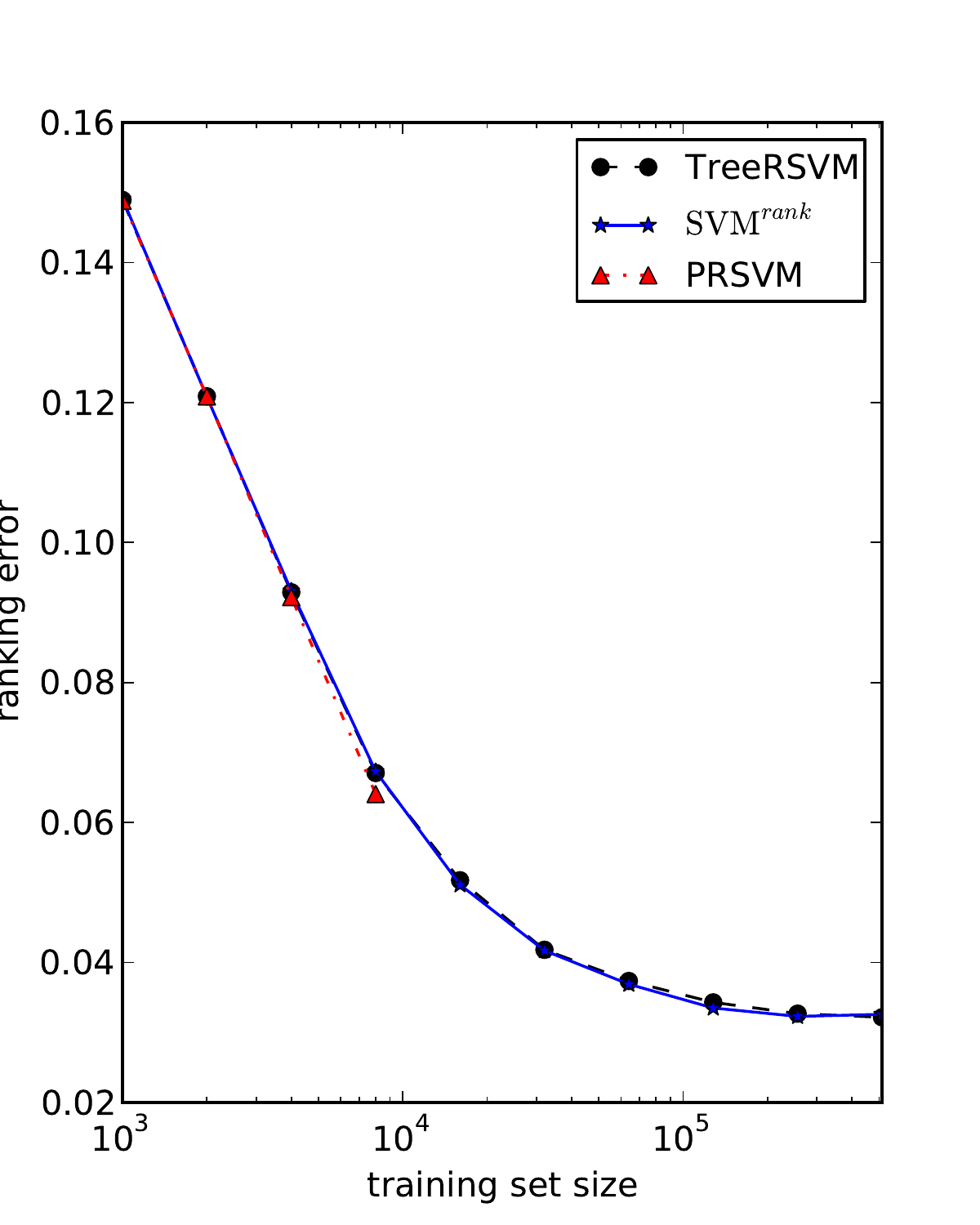}\\
\end{tabular}
\caption{Test error for different RankSVM implementations. Cadata
(left) and Reuters (right).}
\label{fig:perf}
\end{figure}

In the computational experiments we compare the scalability of the
proposed $O(\tsize\nzfsize+\tsize\log(\tsize))$ time training algorithm
to the fastest previously known approach. In addition, we compare our
implementation to the existing publicly available RankSVM solvers.
The considered data sets each contain a single global ranking, and the utility
scores are real valued. This means that $r\approx\tsize$, and the number of
pairwise preferences in the training sets grows quadratically with $\tsize$.
In Section~\ref{setupsection} we describe the experimental setup,
and in Section~\ref{resultsection} we present the experimental results.

\subsection{Experimental setup}\label{setupsection}

We implement the proposed method, denoted as TreeRSVM, as well as
a baseline method PairRSVM, which iterates over all pairs to calculate the
frequencies necessary for the loss and subgradient computation.
Both methods are based on our own implementation of BMRM, and are
integrated to the RLScore\footnote{\url{http://www.tucs.fi/rlscore}}
open source machine learning software framework developed by us.
The majority of the code is written in Python.
All matrix operations are implemented in NumPy and SciPy,
and for solving the quadratic program arising in each BMRM iteration we use
the CVXOPT\footnote{\url{http://abel.ee.ucla.edu/cvxopt/}} 
open source convex optimization software. The most computationally demanding
parts of the subgradient and loss computations, including the search tree
implementation, are written in C language. The only difference between
the two implementations is in the subgradient computation routine.

In addition, we compare our method to the fastest publicly available
previous implementations of RankSVM. The $\mathrm{SVM}^{rank}$ software
is a C-language implementation of the method described by
\citet{joachims2006training}. In theory, $\mathrm{SVM}^{rank}$ and
PairRSVM implement exactly the same method. In practice, implementational
differences such as the use of different quadratic optimizers, and the inclusion
of certain additional heuristics within $\mathrm{SVM}^{rank}$, mean that there
may be some differences in their behavior. PRSVM implements in MATLAB a
truncated Newton optimization based method for training RankSVM
\citep{chapelle2009efficient}. PRSVM optimizes a slightly different objective
function than the other implementations, since it minimizes a squared version
of the pairwise hinge loss. Finally, there exists an implementation
of RankSVM in the $\mathrm{SVM}^{light}$ software package. It has however been
previously shown to be orders of magnitude slower than either
$\mathrm{SVM}^{rank}$ or PRSVM
\citep{joachims2006training,chapelle2009efficient}, and therefore we do not include it in our comparison.

TreeRSVM has $O(\tsize\nzfsize+\tsize\log(\tsize))$ training time
complexity, whereas all the other methods have $O(\tsize\nzfsize+\tsize^2)$
training time complexity. Therefore, TreeRSVM should on large datasets scale
substantially better than the other implementations. Further, all the
methods other than PRSVM have $O(\tsize\nzfsize)$ memory complexity
due to cost of storing the data matrix. PRSVM has $O(\tsize\nzfsize+\tsize^2)$
memory complexity, since it also forms a sparse data matrix that contains two
entries per each pairwise preference in the training set.
\citep{chapelle2009efficient} also describe an improved version of the
method which they state to have similar scalability as
$\mathrm{SVM}^{rank}$, but there is no publicly available implementation
of this method.



The experiments are run on a desktop computer with 2.4 GHz Intel Core 2 Duo
E6600 processor, 8 GB of main memory, and 64-bit Ubuntu Linux 10.10 operating
system. For TreeRSVM, PairRSVM and $\mathrm{SVM}^{rank}$ we use the termination
criterion $\epsilon<0.001$, which is the default setting of
$\mathrm{SVM}^{rank}$. The $\epsilon$ parameter has exactly the same meaning
and scaling for all of these implementations.
For PRSVM we use the
termination criterion Newton decrement $<10^{-6}$, as according to
\citet{chapelle2009efficient} this is roughly equivalent to the termination
criterion we use for the BMRM based methods. $\mathrm{SVM}^{rank}$ and PRSVM
use a regularization parameter $C$ that is multiplied to the empirical risk
term rather than $\lambda$, and do not normalize the empirical risk by
the number of pairwise preferences $\normalizer$. Therefore, the proper
conversion between the $\lambda$ and $C$ values is
$C=\frac{1}{\lambda\normalizer}$. 

We run experiments of two publicly available data sets. Cadata
\footnote{\url{http://www.csie.ntu.edu.tw/~cjlin/libsvmtools/datasets/}}
is a low-dimensional data set consisting of approximately
approximately $20000$ examples, each having 8 features.
The real valued labels are used directly as utility scores. 
Our second data set is constructed from the Reuters RCV1 collection
\citep{lewis2004rcv1}, and consists of approximately $800000$ documents.
Here, we use a high dimensional feature representation, with each
example having approximately $50000$ tf-idf values as features. 
The data set is sparse, meaning that most features are zero-valued.
The utility scores are generated as follows. First, we remove one target example
randomly from the data set. Next, we compute the dot products between each
example and the target example, and use these as utility scores. In effect,
the aim is now to learn to rank documents according to how similar they are
to the target document.

Similarly to the scalability experiments of \citet{chapelle2009efficient},
we compute the running times using a fixed value for the regularization
parameter, and a sequence of exponentially growing training
set sizes. We use $\lambda=10^{-1}$ for Cadata, and
$\lambda=10^{-5}$ for Reuters, as these were observed to lead to good test
performance. The relative differences in running times between the methods were observed to
be similar also for any other tested choices of the regularization parameter
values, though the absolute runtimes are for all the methods the
larger the smaller the value of $\lambda$ is. For Cadata we consider the
training set sizes [1000, 2000, 4000, 8000, 16000]. For Reuters we consider the training set sizes [1000, 2000, 4000, 8000, 16000,
32000, 64000, 128000, 256000, 512000]. 

\subsection{Experimental results}\label{resultsection}

In Figure~\ref{fig:itcost} we plot the average time needed for subgradient
computation by the TreeRSVM and the PairRSVM.
It can be seen that the results are consistent with the computational complexity
analysis, the proposed method scales much better than the one based on
iterating over the pairs of training examples in subgradient and loss
evaluations. On Reuters with half a million training examples PairRSVM
already takes $2760$ seconds (46 minutes) to finish a single iteration, whereas
the same is achieved by TreeRSVM in $7$ seconds.

Next, we compare the scalability of the different RankSVM implementations.
In Figure~\ref{fig:rtimes} we present the runtimes of all the different
implementations, when trained to convergence. As expected, TreeRank achieves
orders of magnitude faster training times than the other alternatives. PRSVM
could not be trained beyond $8000$ examples due to large memory consumption.
On Cadata $\mathrm{SVM}^{rank}$ showed much worse scalability than should be
expected. More detailed study of the $\mathrm{SVM}^{rank}$ results revealed
that almost all of the runtime was consumed by the quadratic solver, which on
some iterations failed to make progress for a substantial amount of time.
The behavior seemed to be caused by numerical problems, our implementations
which use the CVXOPT solver did not have similar issues. On the Reuters data
$\mathrm{SVM}^{rank}$ did not have such problems, the method showed similar
scaling as PairRSVM, as expected. With $512000$ training examples on Reuters
training $\mathrm{SVM}^{rank}$ took $83$ hours, and training PairRSVM took $122$
hours, whereas training TreeRSVM took only $18$ minutes in the same setting.

In Figure~\ref{fig:memuse} we plot the peak memory usage of the considered
implementations to give a rough idea about their scalability in terms of
memory efficiency. We present results only for the Reuters data, since the
Cadata has too few training examples and features per example to allow reliable
benchmarking of the methods which have linear scaling in memory usage.
PairRSVM is left out of the comparison, since it has almost identical memory
consumption as TreeRSVM. PRSVM has quadratic memory complexity with respect to
the training set size, and therefore consumes several GB of memory already at $8000$ training examples.
Both TreeRSVM and $\mathrm{SVM}^{rank}$ start to show linear behavior in memory
complexity once the sample size grows large enough, as expected. At first
the Python based TreeRSVM implementation is much less memory efficient than
the C-language $\mathrm{SVM}^{rank}$ implementation, but as the sample size
grows the difference becomes smaller. For the largest sample sizes TreeRSVM
uses roughly 2.5 times the amount of memory used by $\mathrm{SVM}^{rank}$. The
difference is due to the fact that the TreeRSVM implementation
maintains two copies of the data matrix, one optimized for fast row- and one
for fast column access. Better memory efficiency could thus
be achieved by maintaining only one copy of data matrix, but initial
experiments showed that this would lead to roughly sevenfold increase in
training time as measured on Reuters data.

Finally. in Figure~\ref{fig:perf} we plot the pairwise ranking errors,
for the different implementations, as measured on independent test sets.
For Cadata we use $4000$, and for Reuters $20000$ test examples for computing
the test errors. PairRSVM is left out of the comparison, since it always reaches
exactly the same solution as TreeRSVM. These
measurements act as a sanity check, showing that despite the
implementational differences TreeRSVM and $\mathrm{SVM}^{rank}$ reach
similar performance. Further, we see in the results that even though
PRSVM optimizes a squared version of the pairwise hinge loss, it still
achieves similar test performance as the other methods. 

\section{Conclusion}\label{conclusionsection}

In this work we have proposed an $O(\tsize\nzfsize+\tsize\log(\tsize))$ time
method for training RankSVM. Empirical results support the complexity
analysis, showing that the method scales well to large data sets.
The experiments demonstrate orders of magnitude improvements
in training time on large enough data sets, compared to the
fastest existing previous implementations.
Though we have only considered the linear RankSVM, the approach
could also be used to speed up its kernelized version
using a reduced set approximation, such as the one proposed by
\citet{joachims09sparse}.
A possible future research direction would be to improve the convergence
speed of the BMRM for RankSVM training by devising a line search procedure similar to
the one proposed by \citet{franc2009optimized} for
SVM classification.

\section*{Acknowledgment}
This work has been supported by the Academy of Finland.

\bibliographystyle{elsarticle-harv}
\bibliography{myBibliography}

\end{document}